\documentclass[letterpaper]{article} 
\usepackage{aaai24}  
\usepackage{times}  
\usepackage{helvet}  
\usepackage{courier}  
\usepackage[hyphens]{url}  
\usepackage{graphicx} 
\usepackage{natbib}  
\usepackage{caption} 
\usepackage{booktabs}
\usepackage{newfloat}
\usepackage{amsmath,amssymb}
\usepackage{fancyhdr}

\DeclareMathOperator*{\argmax}{argmax}

\usepackage[noend,ruled,vlined]{algorithm2e}
\usepackage{verbatim}
\usepackage{etoolbox}
\AtBeginEnvironment{algorithm}{\SetArgSty{textrm}}
\SetKwFor{For}{for }{ do}{}
\makeatletter
\patchcmd{\@algocf@start}
  {-1.5em}
  {0pt}
  {}{}
\makeatother

\SetCommentSty{mycommfont}

\DeclareCaptionStyle{ruled}{labelfont=normalfont,labelsep=colon,strut=off} 

\newcommand{\token}[1]{$[$$\scriptstyle\mathsf{#1}$$]$}

\newcommand{\Default}{\text{Def}}

\newcommand{\ClustWP}{\text{Clust-W/P}}
\newcommand{\TopicBased}{\text{TopicBased}}

\newcommand{\RCov}{\text{\textit{RCov}}}
\newcommand{\Cov}{\text{\textit{Cov}}}
\newcommand{\Div}{\text{\textit{Div}}}
\newcommand{\Repr}{\text{\textit{Repr}}}
\newcommand{\CosSim}{\text{\textit{CosSim}}}

\title{Tell Me What Is Good About This Property:\\ Leveraging Reviews For Segment-Personalized Image Collection Summarization}
\author {
    Monika Wysoczanska\textsuperscript{\rm 1},
    Moran Beladev\textsuperscript{\rm 2}, Karen Lastmann Assaraf\textsuperscript{\rm 2}, Fengjun Wang\textsuperscript{\rm 2}, Ofri Kleinfeld\textsuperscript{\rm 2}, Gil Amsalem\textsuperscript{\rm 2}, Hadas Harush Boker\textsuperscript{\rm 2}
}
\affiliations {
    \textsuperscript{\rm 1}Warsaw University of Technology, Warsaw, Poland\\
    \textsuperscript{\rm 2}Booking.com, Tel Aviv, Israel\\
    monika.wysoczanska.dokt@pw.edu.pl
}

\begin{document}

\thispagestyle{fancy}
\chead{This paper was accepted to IAAI 2024. Please reference it instead once published}
\rhead{} 

\maketitle

\begin{abstract}
Image collection summarization techniques aim to present a compact representation of an image gallery through a carefully selected subset of images that captures its semantic content. When it comes to web content, however, the ideal selection can vary based on the user's specific intentions and preferences. This is particularly relevant at Booking.com, where presenting properties and their visual summaries that align with users' expectations is crucial.
To address this challenge, we consider user intentions in the summarization of property visuals by analyzing property reviews and extracting the most significant aspects mentioned by users. 
By incorporating the insights from reviews in our visual summaries, we enhance the summaries by presenting the relevant content to a user. Moreover, we achieve it without the need for costly annotations. Our experiments, including human perceptual studies, demonstrate the superiority of our cross-modal approach, which we coin as \emph{CrossSummarizer} over the no-personalization and image-based clustering baselines.
\end{abstract}

\section{Introduction}

Visual content is one of the key aspects when evaluating and deciding upon a place to stay on the Booking.com platform. Throughout their journey, platform users browse through visual content for four main reasons: (1) To get an accurate and realistic idea of what to expect, (2) To assess the quality of the property, (3) To build trust and remove doubts that they are making the right booking decision, (4) To look for a content that matches their travel intent.
When looking for their next trip, users might be overwhelmed with the amount of information they are exposed to, both visual and textual. Image galleries can contain up to hundreds of images. Hence, we aim to focus the platform's users on the visual content which is the most relevant to them given their current personal context.
We achieve that by summarizing each property with a subset of visually informative, high-quality, segment-personalized images.

\begin{figure}[h!]
    \begin{center}
\includegraphics[trim=0 60 0 0, clip,width=0.5\textwidth]{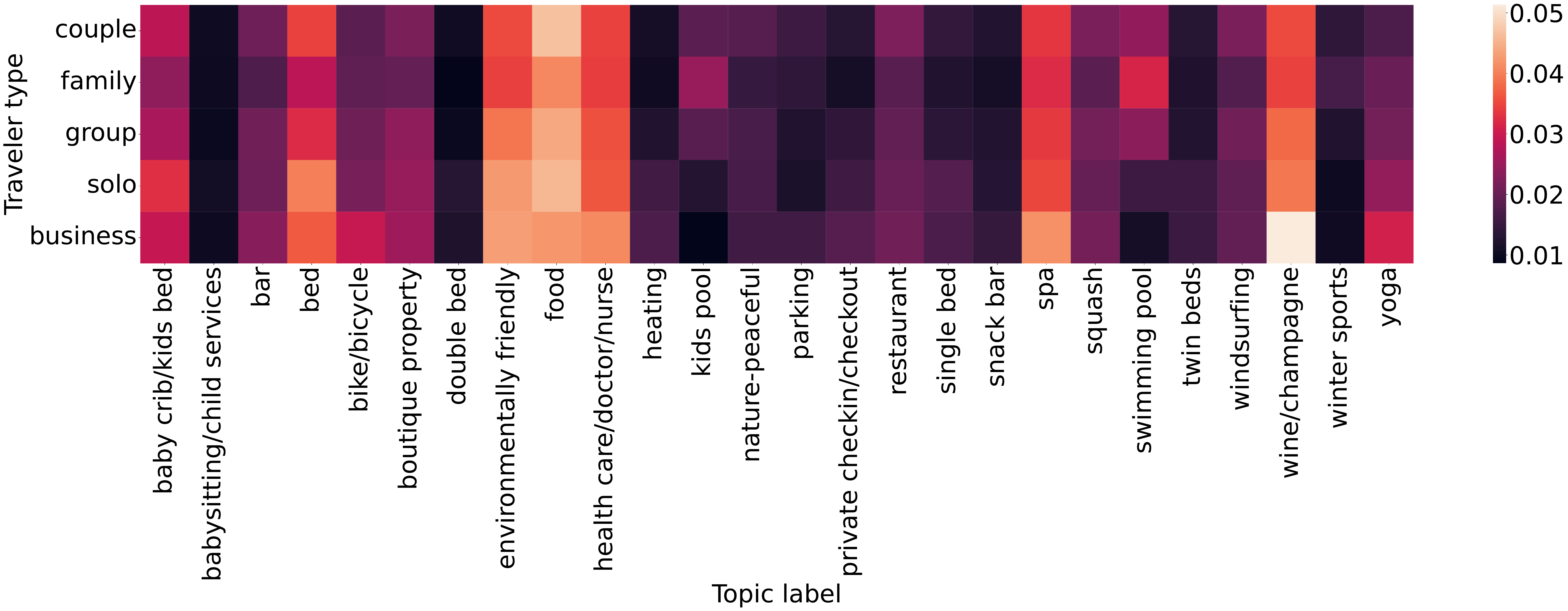}
    \end{center}
    \vspace{-0.36cm}
    \caption{\textbf{Heatmap of most popular topics} (x-axis) extracted from reviews for different traveller types at Booking.com. We note that the ranking of the most mentioned topics differs among traveller segments making reviews a valuable signal for user segment personalization.}
    \label{fig:heatmap}
\end{figure}

Most of the works in the image collection summarization area focus on a generic summarization problem, where the main objective is to select a \emph{diverse} set of images. Only recently, some of the efforts have been made in a so-called \textit{guided summarization}~\cite{Kothawade_Kaushal_Ramakrishnan_Bilmes_Iyer_2022}, where the aim is to get a diverse yet \emph{representative} subset of images corresponding to a specific query. In our work, personalization can be seen as a variant of a query-based approach. However, the queries are not explicit. User intents can not easily be translated into specific queries, making \emph{personalized image gallery summarization} more complex.
In this work, inspired by recent advances in multi-modal learning, we solve the above-mentioned challenge and develop a method for personalized image collection summarization with textual guidance. We focus on entire groups of users, which we further refer to as \textbf{user segments}. We leverage millions of reviews corresponding to properties on the Booking.com platform for enhanced user segment personalization. We do that by extracting key topics mentioned in the reviews, and since they significantly differ across segments of users (see Fig.~\ref{fig:heatmap}) we adjust the image selection accordingly.
As the main challenges of our task, we identify the following:

\paragraph{\textbf{Personalization modelling}} It is not apparent how to obtain the personalization data for our task and avoid costly annotations. Therefore, we focus on the entire segments of users and leverage the textual reviews with the metadata available on the Booking.com platform.

\paragraph{\textbf{User intents extraction from reviews}} Reviews available on the platform can be a very rich source of information. Users of the platform can help future travellers to make the best choices by sharing their experiences. However, in practice, the reviews tend to be very noisy. Therefore, an essential aspect of our approach is to extract relevant pieces of information for a visual summary from free-form text. Thus, the extracted signal has to be adequate for matching the semantic content of images in a gallery.

\paragraph{\textbf{Matching text and images}} Finally, matching text with images requires representing all of them in a joint multi-modal space. There has been a recent surge of methods that leverage free-form text to reduce the need for costly annotations~\cite{narasimhan2022tl, wang2023mumic, DBLP:journals/corr/abs-2007-14937} yielding better performances when cast into multi-modal problems~\cite{DBLP:journals/corr/abs-2201-02494, DBLP:journals/corr/abs-2107-00650, Zhang_Meng_Wang_Jiang_Liu_Yang_2022, zhu-etal-2018-msmo, Zhu_Zhou_Zhang_Li_Zong_Li_2020, visapp20}, or obtaining better image representations~\cite{morgado_avid_cma, DBLP:journals/corr/abs-2103-00020}.
Most importantly, CLIP~\cite{DBLP:journals/corr/abs-2103-00020} shows that image-text large-scale pretraining gives the ability to learn the generalizable image representations and enables zero-shot image-text matching, which we leverage in our approach.

The main contribution of our work is as follows:
\begin{itemize}
    \item We introduce an unsupervised method for image collection summarization personalized for entire segments of users using textual guidance extracted from reviews. Our approach leverages text and image representations in multi-modal space.
    \item We extend previously introduced evaluation metrics for image collection summarization to the segment-personalization use case when no ground-truth annotations are given.
    \item In our experiments, we conduct human perceptual studies alongside the quantitative evaluation using our proposed metrics and show that reviews provide an adequate signal for segment personalization.
\end{itemize}

\begin{figure}[h!]
    \begin{center}        
    \includegraphics[width=0.48\textwidth]{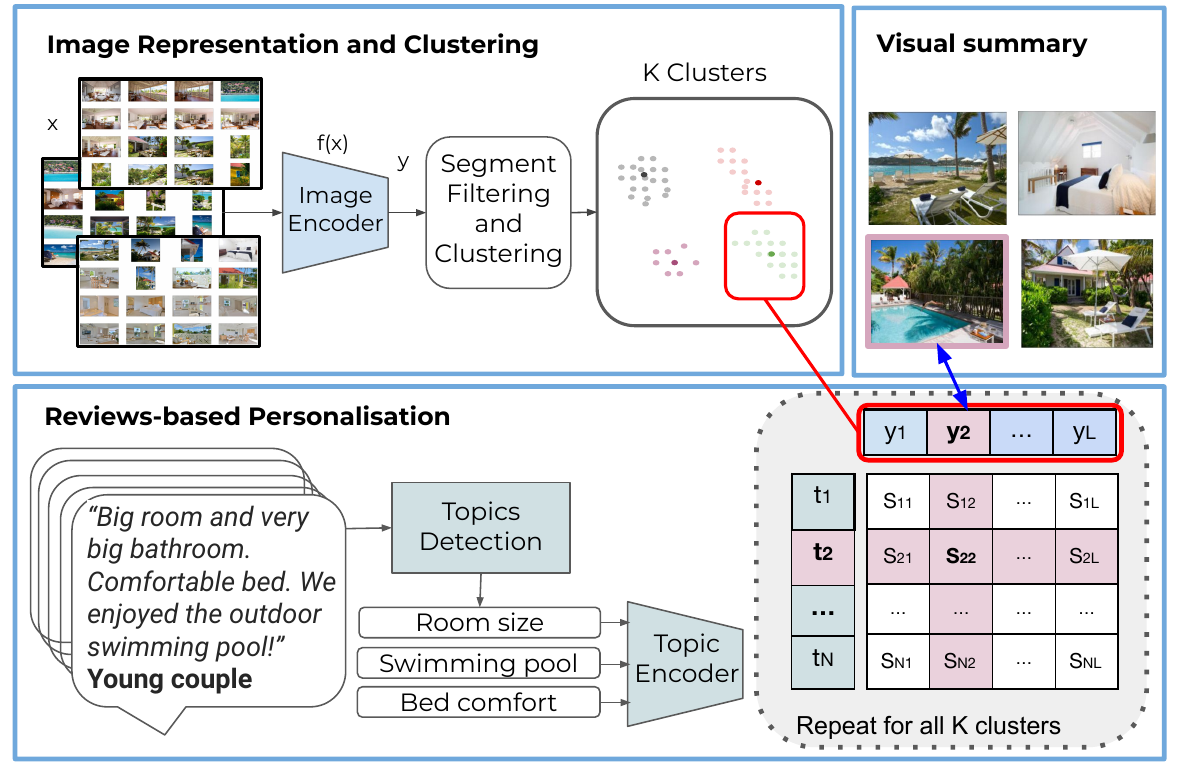}
        \caption{\textbf{Overview of our method}. First, we extract image embeddings and cluster them to obtain K (K=4 in this case) semantically separated groups of images. Then, for each cluster, we calculate the similarities between all the images within the cluster and the topics extracted from reviews of the specific segment $u$ (here u=\textit{Couple}). Finally, the selected images are the ones with the highest similarity to any of the topics.}
        \label{fig:overview}
    \end{center}
\end{figure}

\section{Method} \label{sec:method}
In this section, we define a task of personalized image collection summarization for segments of users and describe in detail all the building blocks of \textit{CrossSummarizer}. We also provide the metrics which we use to evaluate our method quantitatively. We explain how we adapt standard metrics for image collection summarization to our segment-personalized use case.

\subsection{Task Definition}
Given a collection of images $G$, the goal of our method is to select a subset of $K$ images $G_s \subset G$ that best corresponds to a given user segment $u\in U$. 
We consider two types of user segments:
\begin{itemize}
    \item Traveller types: \textit{Solo, Couple, Group, Family, Business}
    \item Trip types: \textit{Beach, Ski, City, Nature Active, Nature Peaceful}
\end{itemize}
Our personalized summary aims to cover the essential aspects of the whole image gallery while selecting images that show details relevant to the specified user segment $u$. 

Each of the images in $G$ has a corresponding set of visual classes $c \in C$, and for each user segment $u$, we manually define a subset of $C_u \subset C$ of relevant classes (for details on how we define $C$ see Appendix~\ref{ap:definition}).
Moreover, we have access to each property's textual reviews, labelled with the corresponding segment $u$. 

However, reviews can mention many aspects in just a short piece of text (see Fig.~\ref{fig:overview}). Therefore, we represent each review as a set of topics $T$ it covers. 

Below we explain all of the steps of our method in detail.  

\subsection{Image Embeddings \& Filtering}
\label{sec:image_embedd_and_filt}
Let $f(x_i) \rightarrow y_i \in \mathbb{R}^D$ be a feature representation of an image $ x_i \in G$. In this work, we use a recently proposed MuMIC~\cite{wang2023mumic} for image representation $f(x)$, which extends the image-text contrastive pretraining proposed in CLIP~\cite{DBLP:journals/corr/abs-2103-00020} to a multi-label case.
MuMIC learns a multi-modal embedding space by jointly training an image encoder and text encoder to maximize the cosine similarity of the image and label-text embeddings of real labels. The model applies tempered sigmoid-based Binary Cross-Entropy (BCE) loss on each class and then mean-reduces it (see Eq. \ref{'f:lnc'}), optimizing across all classes. 
Given a batch of images $\{x_i \in G, i=1..N\}$ and their associated ground-truth multi-label vector $\{\hat{w}_i \in \mathbb{R}^{|C|}, i=1..N\}$
\begin{equation}
\label{'f:lnc'}
\begin{aligned}
\ell_{BCE} = - \frac{1}{N |C|} \sum_{i=1}^{N} \sum_{j=1}^{|C|} ({p_j \hat{w_{ij}}) \cdot \log \sigma(w_{ij})} \\
{+ (1 - \hat{w_{ij}}) \cdot \log (1 - \sigma(w_{ij})))  } 
\end{aligned}
\end{equation}

where $\sigma(\cdot)$ is the Tempered Sigmoid function; 
$w_{ij}$ are the original output logits (before applying temperature scaling), which is the pairwise image-text cosine similarity between image $x_i$ and class $j$; 
and $p_j$ is the positive sample weight of class $j$. 
A higher $p_j$ indicates that positive samples are given greater weight, increasing the penalty for identifying false negatives.

We use MuMIC to both extract image embedding as well as to obtain classes $C$ per each image $x \in G$. 
We then filter out images that are not relevant to given $u$ by keeping only the subset of images with at least one class $c \in C_u$.

\subsection{Clustering}
For the clustering phase we use \textit{K}Medoids algorithm~\cite{PARK20093336} implementation \footnote{\url{https://github.com/scikit-learn-contrib/scikit-learn-extra}}. We choose \textit{K}Medoids as it is robust to outliers and gives high flexibility in choosing \textit{K}.

We run clustering on the image embeddings using \emph{Cosine Similarity} (\CosSim) as a similarity metric, such that

\begin{equation}
\CosSim(u, v)= <\dfrac{u}{\|u\|}, \dfrac{v}{\|v\|}>, \quad \text{with } u, v \in \mathbb{R}^D
\label{eq:cosine_similarity}
\end{equation}

where $<\cdot,\cdot>$ is the inner product in $\mathbb{R}^D$.

Finally, we obtain resulting cluster assignments for each of the images. 

\subsection{Text2topic: Topics Detection Model}
We use a recently proposed Text2Topic~\cite{wang2023text2topic} model, a topic detection model, to extract user segment preferences and personalize the subset of images based on their topics of interest.
We first filter the reviews by $u$ and then detect the topics associated with these reviews.
To get the topics, we train a classification model with 45 travel-domain topics (see the subset of the topics in the heatmap in Fig.~\ref{fig:heatmap}) using cross-encoder transformer-based architecture \cite{reimers-2019-sentence-bert}, which relies on BERT~\cite{devlin-etal-2019-bert}. We train the model with 15,663 positive pairs of reviews and topics and sample X5 negative topics per review. TWe use the Binary Cross Entropy loss function on the \token{CLS} embedding vector of the crossed input \token{REVIEW} \token{SEP} \token{TOPIC} to get the probability the topic is mentioned in the review. 

At inference, we run the model on pairs of reviews and each of the 45 topics to get the probability scores. Finally, we get the topics $T$ with a probability greater than 0.5 to match with the review.

\subsection{Matching Images to Topics}
Having obtained clusters of images and a list of topics $T$ for given $u$, $T_u \subset T$ we select the final subset of images $G_s$. We do it by first computing the \emph{confidence matrix $S \in \mathbb{R}^2$} of images $G$ being aligned with topics $T_u$. We follow~\cite{wang2023mumic} and use tempered sigmoid, formulated as:
\begin{flalign}
\begin{split}
S_{ij}
&= \sigma(\text{exp}(\gamma)\cdot <t_{i},y_{j}>)
\end{split}
\label{eq:confidence}
\end{flalign}  

where $\gamma$ is the log-parameterized multiplicative scalar, and $t_i, y_j\in \mathbb{R}^D$ are respectively a topic from $T_u$ and a feature representation $f(x_j)$ of an image $x_j$.

We then select the final images by iterating over clusters and selecting the pair $(t_i, y_j)$ with the highest similarity within a cluster of representations. The pseudo-code for this selection is given in the Algorithm~\ref{algo:pseudocode}.

\begin{algorithm}[t]
\SetKwInOut{Input}{Input}
\SetKwProg{Init}{Initialize}{}{}
\SetAlgoLined
\SetKwFunction{selectImages}{\text{SelectImages}}
\SetKwProg{myproc}{procedure}{}{}
\KwResult{$G_s$ - selected images from the gallery}
\Input{}
$A_K$ - Image-to-cluster assignment : $x \in G \mapsto \{1..K\}$

$S$ - Confidence matrix from Eq.~\eqref{eq:confidence}

\myproc{\selectImages{$L_{k}$, $S$}}{
  $G_s \gets \emptyset$  \tcp*[f]{Selected images}
  
  $\Omega_T \gets \{1..|T|\}$ \tcp*[f]{Indices of active topics} 

  
  \For{Cluster $k \in \{1..K\}$}{
    \tcp{Image indices for cluster k} 
    
    $\Omega_k \gets \{ j : A_K(x_j) = k, \>\> \forall j \in \{1..|G|\}\}$ 
    ~\\
    ~\\
    
    \tcp{Compute best matches within $k$}
    $i^*, j^* \gets \argmax\limits_{(i, j) \in \Omega_T \times \Omega_k}  S_{ij}$ 
    ~\\
    ~\\
    
    
    $\Omega_T \gets \Omega_T \setminus \{i^*\}$  \tcp*[f]{Update active topics}
    
    $G_s \gets G_s \cup \{x_j^*\}$  \tcp*[f]{Add selected image}
  }
 \KwRet $G_s$
 }{}
\caption{Pseudocode for selecting images}
\label{algo:pseudocode}
\end{algorithm}






\subsection{Evaluation Metrics}
To ensure our generated summaries are diverse, and adequately correspond to user segments' interests, we define multiple evaluation metrics. Following~\cite{DBLP:journals/corr/abs-1809-08846}, we use \textit{Coverage, Representativeness} and \textit{Diversity} metrics. However, we apply some modifications to match our use case. More specifically, we make sure each one of the metrics is normalized across samples. Our motivation lies in a large variety of galleries across properties regarding image redundancy and coverage of relevant user segment aspects within images. 

\paragraph{Diversity} Let us denote $d(x_i, x_j)$ as a distance between images $x_i$ and $x_j$. The \textit{Diversity} (\Div)  metric measures to what extent the diversity in terms of distance between embeddings of images in $G_s$ is similar to the one in the original gallery $G$. We define Div as:
\begin{equation}
\Div = \dfrac{\max\limits_{(x_i, x_j) \in G_s \times G_s } d(x_i, x_j)}{\max\limits_{(x_l, x_m) \in G \times G} d(x_l, x_m)}, 
\end{equation}
where $d(\cdot, \cdot)$ is computed in image embedding space as:

\begin{equation}
    d(x_i, x_j) = 1 - \CosSim(f(x_i), f(x_j)).
\end{equation}

\begin{table*}[h]
\centering
\begin{tabular}{cccccc}
\toprule
 Split & No. reviews & No. images &  No. samples & Total No. reviews & Total No. images \\
 \midrule
Small &  between 30 and 150 &  between 50 and 100 & 3230 & 251943 & 232560 \\
 Big &  151 $<$ &  100 $<$ &   3151 &  1504024 & 457946 \\
 \bottomrule
\end{tabular}
 \vspace{0.2cm}
 \caption{Dataset split in detail. Overall we collected more than 6000 samples from the platform. Through stratified sampling, we make sure the distribution in terms of location, rating, and accommodation type reflects the real distribution.}
  \label{tab:ds_splits}
\end{table*}

\paragraph{Representativeness} 

Then, let us denote $\mu_{G}, \mu_{G_s} \in \mathbb{R}^D$
as the mean vectors of the original gallery representations and the selected subset, respectively. \textit{Representativeness} (\Repr) is defined as:
\begin{equation}
\Repr = \CosSim(\mu_{G},\mu_{G_s})
\end{equation}
With the generated summaries, we wish both vectors to be similar in representation, such that their CosineSimilarity is close to 1.

\paragraph{Coverage} We also measure \textit{Coverage} (\Cov) in the semantic space by using classes associated with images in $G$. We adopt the \textit{Probabilistic coverage} suggested in~\cite{DBLP:journals/corr/abs-1809-08846} and use the probabilities $P(c|x) \in \mathbb{R}$ for a given image $x$ to represent a particular class $c \in C_u$.

\begin{equation}
\Cov = \dfrac{1}{|C_u|} \sum_{c \in C_u} \dfrac{P_{G_s}(c)} {P_G(c)}, 
\end{equation}

where $P_G(c) = \max_{x\in G} P(c|x)$.
Note that in our work we use MuMIC model to obtain probabilities since it was trained in-domain. However, this could be any off-the-shelf multi-label image classification method. \\
With our \textit{Coverage} metric, we mainly focus on measuring the accuracy of the personalization step. Therefore, instead of taking all of the classes in $C$, we only consider the ones corresponding to a given user segment $u$, $C_u$. 

\paragraph{Reviews Coverage} Finally, to measure how well the generated summaries correspond to user segment topics from reviews, we calculate the topics coverage of selected images. Similarly to \textit{Coverage}, we use the confidence matrix $S \in \mathbb{R}^2$ as probabilities of a topic $t_j$ being aligned with the set of selected images $G_s$. \textit{Reviews Coverage} (\RCov) is therefore given by: 

\begin{equation} \label{eq:1}
\RCov = \dfrac{1}{|T_u|} \sum\limits_{i=1}^{|T_u|}\dfrac{\max\limits_{j \in \Omega_{G_s}} S_{ij}} {\max\limits_{j \in \Omega} S_{ij}}, 
\end{equation}

where $\Omega = \{1..|G|\}$ and $\Omega_{G_s} = \{j: x_j \in G_s, \forall j \in \Omega\}$ are respectively the range of indices of the images in $G$ and the subset of indices of images that are only in $G_s \subset G$.

\section{Experiments}
\label{sec:experiments}
In this section, we provide the experimental results obtained through both offline evaluation and user studies conducted internally at Booking.com. We also describe our experimental setup, including details on the dataset collected at the Booking.com platform and baseline models we compare our CrossSummarizer against.

\subsection{Experimental Setup}
\label{sec:exp_set}

\subsubsection{Dataset} We conduct our experiments on the Booking.com dataset consisting of properties. We collected over 6000 real properties from the platform by carefully curating the sampling to adequately represent a real distribution in terms of geographical location, types of accommodation as well as travellers' experience with a given property. We do it by applying a stratified sampling technique with a country, type of accommodation, and property rating being the factors. 
We do not share our dataset, however, we note that the examples can easily be downloaded (both images and reviews) since the data is publicly available on the Booking.com site.

Each of the samples in our dataset consists of a set of uploaded by property owners images, which correspond to a gallery, and a set of reviews of past travellers' experiences. Alongside sampled reviews, we also include metadata about the type of traveller that authored a particular review.

Since samples in our dataset significantly vary in amounts of both images and reviews, we split the dataset into two groups according to the size of galleries and the number of reviews. Precisely:
\begin{itemize}
    \item \textbf{Small} - properties with the size of a gallery between 50 and 100 photos and 30 - 150 reviews, 
    \item  \textbf{Big} - gallery sizes of 100 $<$ photos and 150 $<$ reviews.
\end{itemize}
More details on precise numbers of our dataset split are given in Tab.~\ref{tab:ds_splits}. In our experiment, we report results separately for the two aforementioned splits.

\subsubsection{Baselines} To the best of our knowledge, none of the proposed methods for \emph{image collection summarization} such as~\cite{Kothawade_Kaushal_Ramakrishnan_Bilmes_Iyer_2022,NIPS2014_a8e864d0} nor the \emph{multimodal} ones~\cite{Zhu_Zhou_Zhang_Li_Zong_Li_2020,Zhang_Meng_Wang_Jiang_Liu_Yang_2022,ijcai2018p577} tackle the \emph{personalization case}. For evaluation purposes, we implement multiple baselines, which we describe in detail below.

\paragraph{\textit{Default Clustering} (\Default)}
We compare against a simple no-personalization approach by running clustering on image embeddings and omitting the filtering step. The selected \textit{K} images are resulting cluster centres. A similar approach was proposed in~\cite{10.1145/1141277.1141601} for videos.

\paragraph{\textit{Clustering with personalization} (\ClustWP)} We implement the approach without topic-based refinement and select cluster centres as the summarization. The approach differs from the \textit{Clustering} setting by an additional filtering phase based on relevance to the particular user segment classes being associated with images in the gallery (see Sec.~\ref{sec:image_embedd_and_filt}).

\paragraph{\textit{Topic-based personalization} (\TopicBased)} We also implement the approach based only on the topic's similarity with images, without a clustering step. We extract image and topic embeddings and apply user segment filtering. We calculate the similarity matrix between all the image and topic embeddings and choose top \textit{K} scores in the matrix. We make sure each image gets selected only once and apply the selection process iteratively. The pseudo-code is similar to the final approach, as we simply take $\argmax$ on the entire confidence matrix $S$, which is being decreased with each iteration by selected images' columns.  

Note that for all of the approaches, we use the same image representations obtained with MuMIC, which was trained on Booking.com multi-label classification dataset as described in~\cite{wang2023mumic}.
For the filtering phase, we create a mapping of MuMIC's classes for each user segment beforehand and use the mapping at inference time.

\subsection{Offline Evaluation}
\begin{table*}[h]
\centering
\resizebox{0.98\textwidth}{!}{%

  \begin{tabular}{ccccccccc}
\toprule

    & \multicolumn{4}{c}{Small} & \multicolumn{4}{c}{Big} \\
    \midrule
    Method&\Div$\uparrow$ & \Repr$\uparrow$&\Cov$\uparrow$&\RCov$\uparrow$&\Div$\uparrow$&\Repr$\uparrow$&\Cov$\uparrow$&\RCov$\uparrow$ \\
    \midrule

    \multicolumn{1}{l}{\textit{No Personalization}}\vspace{3pt}\\

    Default Clustering~\cite{10.1145/1141277.1141601}& 0.947 & \textbf{0.931} & 0.516 & 0.581 & 0.925 & \textbf{0.929} & 0.430 & 0.488 \\
    \midrule
    \multicolumn{1}{l}{\textit{Personalization}} 
\\
    Clustering with Personalization & 0.903 & \textbf{0.931} & 0.578 & 0.656 & 0.932 & 0.928 & 0.487 & 0.521\\
    Topic-Based Personalization & 0.903 & 0.733 & 0.365 & 0.649 & 0.874 & 0.775 & 0.261 & \textbf{0.707}\\
    CrossSummarizer (Cross-modal)& \textbf{0.963} & 0.903 & \textbf{0.617} & 
\textbf{0.729} & \textbf{0.950} & 0.885 & \textbf{0.524} & 0.677  \\
  \bottomrule
\end{tabular}}
\caption{Quantitative evaluation. We report the results for two different dataset splits: small and big galleries and the number of reviews.}
\vspace{-0.4cm}

\label{tab:results}
\end{table*}

\subsubsection{Evaluation protocol} We first run the evaluation procedure offline internally at Booking.com. For all the methods relying on \textit{K}Medoids algorithm, we set the same random seed for a fair comparison. We run experiments with a fixed $K=9$ corresponding to the current Booking.com setting.  

\subsubsection{Quantitative results}
The results of our experiments are presented in Tab.~\ref{tab:results}.  
Looking at \Cov~and \RCov, we observe that our cross-modal approach outperforms the baselines in the segment personalization task, especially for the \textbf{Small} dataset split. Compared to the \Default~setting, we observe approximately 0.1 gains in 
\Cov~for both dataset splits and a significant improvement over \RCov. For the \textbf{Big} dataset split, we report a higher \RCov~for the~\TopicBased~approach. This is expected since the approach is based on maximizing the similarity between topics and images, which corresponds to the Eq.~\ref{eq:1}. 

Moreover, our cross-modal approach outperforms the rest of the methods in terms of \textit{Diversity}, indicating that the personalization step produces more diverse summarization. 

Relying only on reviews, which corresponds to the results of a Topic-based method, gives a significantly lower \textit{Representativeness} and poor visual user segment \textit{Coverage}. 
\textit{Coverage}~gain of our CrossSummarizer over the \ClustWP~approach also emphasizes the need for using two modalities in our task. Overall, we observe a clear trade-off between \textit{Diversity}~ and \textit{Representativeness}. CrossSummarizer finds a sweet spot between the two, giving an excellent segment personalization result at the same time. 

\subsubsection{Qualitative results}
Alongside the quantitative evaluation, we also provide some qualitative results. Fig.~\ref{fig:applications} shows the visual comparison between the two baseline approaches and our method. The example shows a non-personalized \Default~result and the personalized results obtained with \ClustWP~and CrossSummarizer approach for \textit{Ski} trip type and $K=8$. 

All of the approaches give a set of diverse images without any redundant photos. However, our cross-modal model provides the best personalization result. We highlight the photos that are relevant to the user segment: a picture of a skier and a photo of a sauna. We can also see that contrary to the \ClustWP~approach, with our CrossSummarizer, the selected image of a property from outside (bottom right) was taken in the wintertime (expected for \textit{Ski} trip type).

\subsection{User Studies}

\subsubsection{Experimental setup}
In addition to offline evaluation, we conduct human perceptual studies in the form of an anonymized paired test. We compare two methods: \ClustWP~approach and CrossSummarizer. We manually select 210 samples of properties with galleries that contain photos relevant to the specified user segment. We split the samples across 5 participants. The samples are uniformly distributed over the dataset split (\textbf{Small}, \textbf{Big}) and user segment types. 

The participants were asked to answer the question:~\textit{Which of the models (A or B) gives a better summary for a given user segment?} We assign a score of 1 for a model that performs better than the other and 0 otherwise. We also allow for a 0.5 score in case of a tie. 

\begin{figure}
    \begin{center}
\includegraphics[trim=0 40 0 5, clip, width=0.45\textwidth]{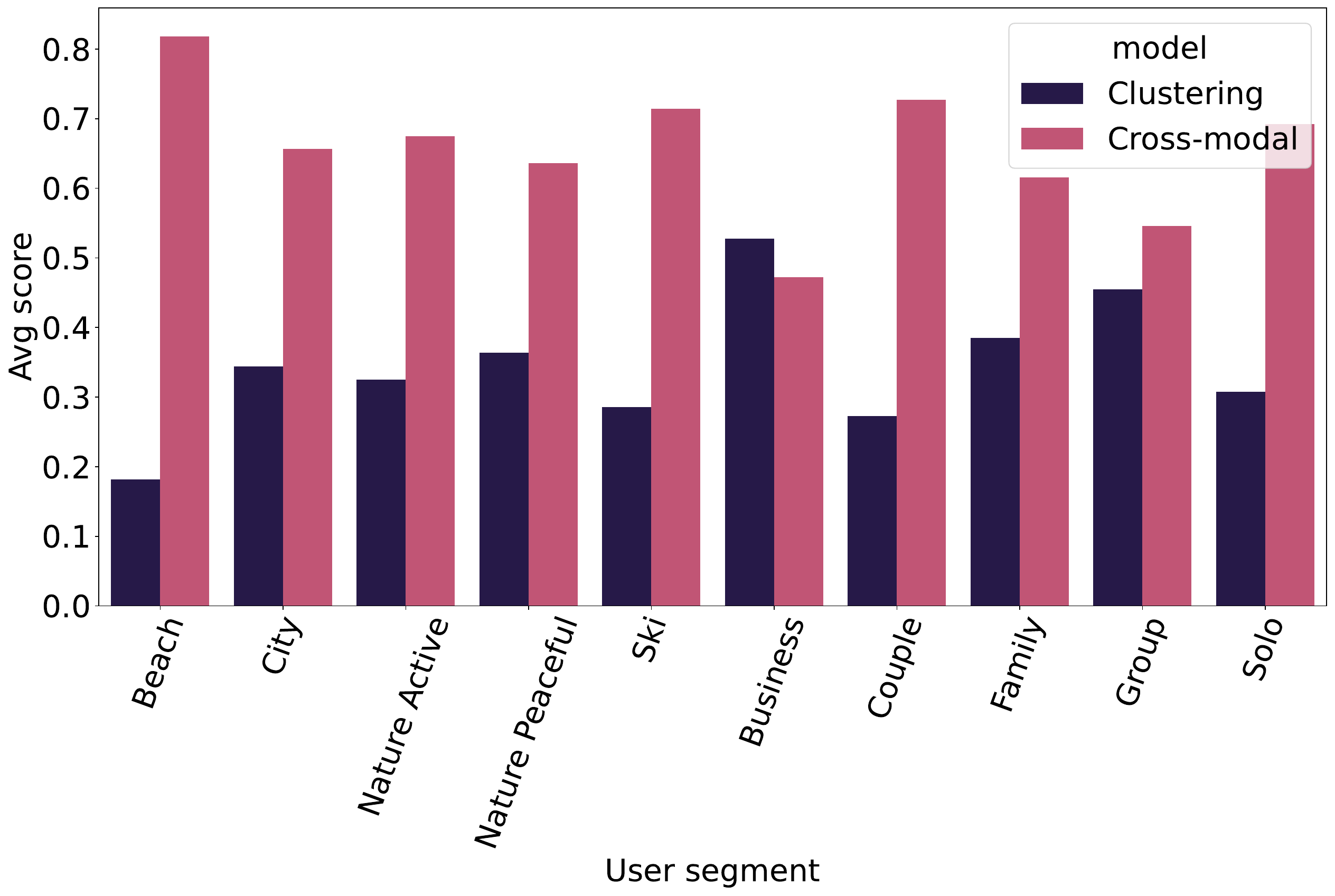}
    \vspace{-0.24cm}
    \caption{\textbf{Results of our user studies.} We report the results separately for each user segment (x-axis).}
    \label{fig:userstudy}
    \end{center}
\end{figure}

\subsubsection{Results} The conducted user studies indicate the superiority of our CrossSummarizer approach with an average score of 
$0.66 \pm 0.38$ for all user segments over \ClustWP~$0.34 \pm 0.38$. Additionally, in Fig.~\ref{fig:userstudy}, we provide detailed results for each user segment. We observe that CrossSummarizer obtains higher average scores for most of the segments in our experiments, except for only \textit{Business} traveller type. 
To confirm the statistical significance of our studies, we perform the paired T-test~\cite{student1908probable} with a null hypothesis of \emph{average scores for models A and B being equal} and the alternative hypothesis of \emph{model A scoring lower than model B}. The results $(t(209)=-5.537; p<5e-8)$ let us reject the null hypothesis and accept the alternative hypothesis.

\section{Application}
\label{sec:application}

This section covers some practical aspects of our approach and how it is leveraged at Booking.com.
Two essential parts are first run offline, which are the Text2Topic model for topic extraction and the MuMIC model for image embedding extraction and image multi-class annotations. The results are then stored in the database and accessed at runtime. It takes approximately 170 ms for the MuMIC model to run on a batch of 100 images and 244 ms for the Text2Topic model to run prediction on a batch of 100 reviews. We leverage GPU computation for this purpose.

\subsection{Deployment \& Maintenance}
The model is served with Amazon SageMaker and deployed on the Booking.com Content Intelligence Platform (CIP)~\cite{wang2023mumic}. CIP is a stream processing platform based on Apache Flink. It consumes real-time events from Kafka topics (e.g. images uploaded by Booking.com partners) and generates model-based predictions. 
The same architectural design allows CIP to be also used for backfilling purposes.
Backfilling refers to the enrichment of historical data with newly deployed model predictions.
We leverage the mentioned design by simulating events of historical data and pushing them to Kafka.

CIP is designed to achieve high prediction throughput while keeping a low latency.
It achieves that by leveraging Apache Flink's asynchronous I/O operator 
to perform concurrent asynchronous HTTP calls to a model endpoint. However, this optimization mechanism relies on the assumption that each model prediction can be made independently.
This assumption does not hold for summarization models where a group of events should be sent to the model together in a single prediction query.

We built upon Apache Flink's windowing mechanism to implement the grouping of events that should be sent to the model endpoint.
Whenever a new call to the model should be triggered, an event containing the request metadata (e.g. a hotel id) is sent to Kafka. Then, the matching image data is fetched by issuing calls to an external service holding image data. Those calls are executed independently and concurrently.
Following that, images are grouped in the same window. Images are accumulated within the window as soon as they are fetched, and the window is closed after a predefined period (e.g. 3 seconds). Then, the images are sent together to the model endpoint for prediction.

\subsection{Application example} 

\begin{figure}[h]
    \begin{center}
\includegraphics[width=0.47\textwidth, trim = 0 27 0 23, clip]{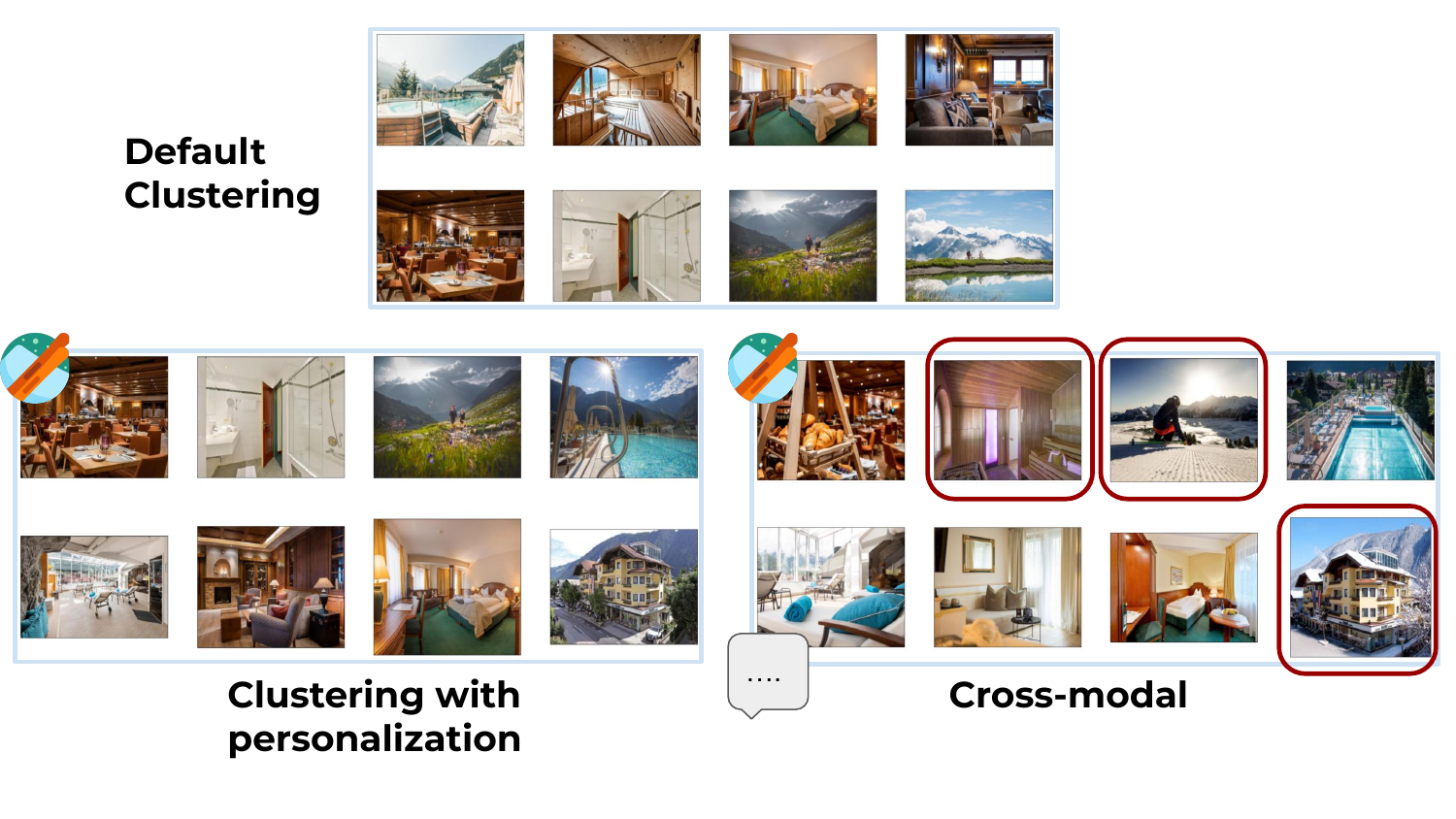}
    \end{center}
    \caption{Example results of summaries of $K=8$ images for one of the properties. We compare Def (top), and personalized methods (bottom) on this property's visual summarization. The presented personalization examples here are for a \textit{Ski} trip type. We highlight in red the selected images which are relevant to this user segment.}
    \vspace{-0.4cm}
    \label{fig:applications}
\end{figure}
 
The applications of our image collection summarization approach at Booking.com are three-fold:
\begin{itemize}
    \item Image subset selection optimization for large collections of images when given a constraint on a number of images or smaller displays.
    \item Visual content personalization based on traveller type.
    \item Visual content personalization based on trip type.
\end{itemize}
Fig.~\ref{fig:applications} shows some qualitative results of our CrossSummarizer for the third bullet point.
Using this personalization approach, we expect to reduce the friction from the decision-making process as users will have a better understanding of the property at an earlier step.

 Our proposed model is currently under experimation. When deploying the personalized CrossSummarizer model we compare it with the current model, which is produced by the \Default~ model, through A/B test experimentation on CTR (click to ratio) metric.

We also note that personalizing summaries of collections of images is a relevant task for most e-commerce websites. Hence, our method could also be applied to any other personalization task, such as product recommendations where a multi-modal input is available (reviews + images). 

\section{Conclusions \& Limitations}
\label{sec:conclusion}
We presented a method for personalized image collection summarization for entire segments of users. Our approach is capable of taking into account users' intents when producing summaries of large image collections. As the personalization signal, we use other travellers' experiences with properties, which we extract from the reviews. We implemented and tested our method on the Booking.com platform and our experiments, including human perceptual study, indicate that our proposed approach yields good results on the diversity and representativeness axes. The comparison with other baselines indicates that our proposed method performs the best in terms of personalization. Future works include A/B tests of our model in a production environment to measure the real-world impact.

The main limitation of our method is handling samples that are considered a \textit{cold-start} zone, e.g. having a limited number of reviews, but complete image galleries. This, however, can be addressed by leveraging information from other properties of a similar profile.

\section*{Acknowledgments} 

The work was supported by Booking.com. Monika Wysoczanska was partially supported by the National Centre of Science (Poland) Grant No.2022/45/B/ST6/02817. We want to thank David Konopnicki, Manos Stergiadis, Sergei Krutikov, Satendra Kumar, and Michael Ramamonjisoa for their feedback. 

\bibliography{aaai24}

\clearpage
\appendix
\section{Appendix}
\subsection{Definining sets of relevant image classes}
\label{ap:definition}
We provide more details on how we defined visual classes for each of the user segment considered in this work. 
We first obtain a list of relevant classes in the Booking.com context from domain experts, i.e. product managers who, based on the statistics of the content of image galleries, decided upon the most frequent classes, leading to a dataset with 120 classes. Having obtained the list of visual classes, we then ask 3 independent experts to associate classes with users. In Tab.~\ref{tab:anns} we present exemplary classes for each considered segment.
\begin{table}[!ht]
\centering
\resizebox{1\columnwidth}{!}{
\begin{tabular}{cc}
\toprule
User Segment    & Image classes      \\      
\midrule
Solo            & Bar, Entertainment Center, Lobby or reception, TV/Multimedia, Public transport                       \\
Couple          & Private dining area, Bed, Sea view, Fireplace, Bathtub, Jaccuzzi/Hot tub                             \\ 
Family          & Aqua park, BBQ facilities, Children playground (outdoors), Dishwasher, Game room                     \\ 
Group           & Buffet, Kitchen or kitchenette, Billiard, Water sport, Seating area (sofa, living room, etc.)        \\ 
Business        & Business traveler, Business/Conference Room, Desk (for work), Lobby or reception, Ironing facilities \\ 
Beach           & Beach, Sea view, Sun umbrella, Sunbed, Sunset                                                        \\ 
Ski             & Heating (not air conditioner), Mountain view, Winter, Fireplace, Skiing                              \\ 
Nature Peaceful & Mountain view, River/Lake view, Natural landscape, Yoga, Garden/garden view                          \\ 
Nature Active   & Yoga, Cycling/Biking, Horse-riding, Birds eye view, Water sport                                      \\ 
City            & Landmark (attractions, sightseeing), Neighborhood/Street, City view, Car, Public transport           \\ 
 \bottomrule

\end{tabular}
}
\caption{Exemplary image classes per user segment.}

\label{tab:anns}

\end{table}

\end{document}